\documentclass[10pt, a4paper]{article}

\usepackage{lrec-coling2024} 
\usepackage{algorithm}
\usepackage{algorithmic}
\usepackage{amsmath}
\usepackage{amsfonts}
\usepackage{booktabs}
\usepackage{tabularx}
\usepackage{caption}
\usepackage{subcaption}
\usepackage{multirow}
\usepackage{enumitem}
\usepackage{tasks}
\usepackage{pifont}

\name{Zichen Wu, Hsiu-Yuan Huang, Fanyi Qu and Yunfang Wu*\thanks{*\ \  Corresponding author.}}

\address{National Key Laboratory for Multimedia Information Processing, Peking University, China \\
School of Computer Science, Peking University, China\\
wuzichen@pku.edu.cn,  wuyf@pku.edu.cn\\}

\def\qqquad{\hskip3em\relax}
\makeatletter
\newcommand*\tablefontsize{%
  \@setfontsize\tablefontsize{8.5}{10.0}%
}
\makeatother

\title{Mixture-of-Prompt-Experts for Multi-modal Semantic Understanding}

\abstract{
Deep multi-modal semantic understanding that goes beyond the mere superficial content relation mining has received increasing attention in the realm of artificial intelligence. The challenges of collecting and annotating high-quality multi-modal data have underscored the significance of few-shot learning. In this paper, we focus on two critical tasks under this context: few-shot multi-modal sarcasm detection (MSD) and multi-modal sentiment analysis (MSA). To address them, we propose Mixture-of-Prompt-Experts with Block-Aware Prompt Fusion (MoPE-BAF), a novel multi-modal soft prompt framework based on the unified vision-language model (VLM). Specifically, we design three soft prompt experts: a text prompt and an image prompt that extract modality-specific features to enrich the single-modal representation and a unified prompt to assist multi-modal interaction.
Additionally, we reorganize Transformer layers into several blocks and introduce cross-modal prompt attention between adjacent blocks, which smoothens the transition from single-modal representation to multi-modal fusion. On both MSD and MSA datasets in few-shot settings, our proposed model not only surpasses the 8.2B model InstructBLIP with merely 2\% parameters (150M), but also significantly outperforms other widely-used prompt methods on VLMs or task-specific methods.
 \\ \newline 
 }

\begin{document}

\maketitleabstract

\section{Introduction}
Multi-modal semantic understanding (MSU) is crucial for the development of machines capable of interpreting the complex interplay of textual and visual information. In social media platforms, where the combination of text and imagery can often present conflicting messages or nuanced sentiments that are not immediately apparent from a single modality alone, such understanding is vital for accurately interpreting the intent and sentiment. Among the fields of MSU, Multi-modal Sarcasm Detection (MSD) and Multi-modal Sentiment Analysis (MSA) emerge as two representing tasks. These tasks exemplify the intricate process of aligning and comprehending the relations from different modalities to discern the intended meaning or sentiment. As highlighted by the examples in Table 1, solely reading the text or image of MSD is prone to misinterpreting it as a positive comment while ignoring the sarcasm about the discount being too small.


Recent approaches for MSU normally exploit a dual-encoder architecture, i.e., use separate pre-trained encoders to extract features for different modalities (e.g., BERT \citep{devlin-etal-2019-bert} for text and ResNet \citep{He_2016_CVPR} for image). The features then interact with each other to capture the incongruity and fed into a classification head for the prediction. 
Under this framework, researchers are dedicated to designing effective methods of interaction, including attention mechanisms \cite{pan-etal-2020-modeling, xu-etal-2020-reasoning, 10.1145/3462244.3479919}, graph structures \cite{10.1145/3474085.3475190, liang-etal-2022-multi, liu-etal-2022-towards-multi-modal}, optimal transport \cite{Pramanick_2022_WACV} and dynamic routing \cite{tian-etal-2023-dynamic}.

\begin{table}[t]
\centering
\scriptsize
\begin{tabular}{@{}c|c|c@{}}
\toprule
Task & MSD & MSA \\ \midrule
Image & 
\begin{minipage}{.12\textwidth}
    \includegraphics[width=0.9\linewidth]{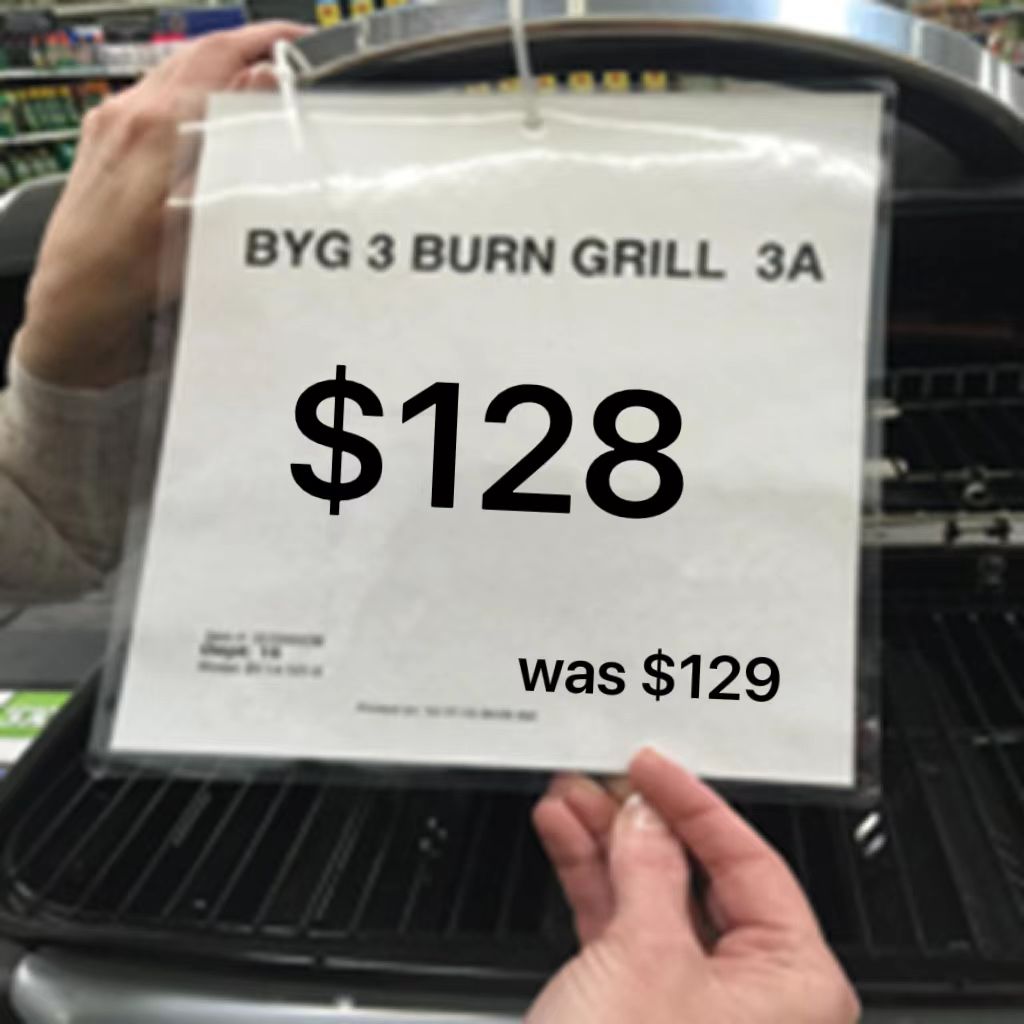}
\end{minipage} &
\begin{minipage}{.12\textwidth}
    \includegraphics[width=0.9\linewidth]{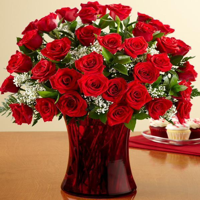}
\end{minipage}\\ \midrule
\multirow{2}{*}{Text} & \multirow{2}{*}{Great Sale!} & \multicolumn{1}{p{0.12\textwidth}}{Wish you a good Valentines Day.} \\ \bottomrule
\end{tabular}
\caption{Examples of MSD and MSA task. The left example displays \textit{sarcasm}, as the text ``great sale'' contradicts with the image depicting a mere 1-dollar discount. The right example conveys a positive wishing attitude, thus categorized as \textit{positive}.}
\label{tab:msd-example}
\end{table}

Although the previous studies have achieved good performance in semantic understanding tasks like MSD or MSA, they mostly rely on sufficient training data. However, collecting a large amount of high-quality multi-modal data, sarcasm especially, is a non-trivial task. According to \citet{MISRA202313}, sarcasm expressions require a high level of cognitive ability and rarely appear on social media, making it difficult to collect and annotate. Moreover, most existing models use separate pre-trained encoders to process text and image, which might lead to the misalignment of different modalities and thus hurt modal fusion. 
Nowadays, pre-trained on large-scale image-text pairs, the vision-language models (VLMs) achieve good image-text correspondences and can perform well on cross-modal reasoning. Given these two aspects, we propose to address the few-shot MSU tasks by using VLMs.

To adapt the pre-trained VLMs to downstream multi-modal tasks on few-shot settings, prompt-based learning is widely applied and has demonstrated promising performance~\cite{10.1145/3560815}. Compared to manually designed prompts, continuous prompts (also referred to as soft prompts) are preferred due to their flexibility and scalability ~\cite{DBLP:journals/corr/abs-2110-07602,liu2021gpt, han2021ptr}. However, most previous work only coped with text data.  
In this paper, we explore different methods of utilizing soft prompts 
to address the few-shot MSU task.


Two primary architectures are prevalent in VLMs.
One line of work 
encodes images and texts respectively and performs modal fusion by simply computing the similarities between them, like CLIP \cite{pmlr-v139-radford21a} and ALIGN \cite{pmlr-v139-jia21b}. 
Another line of work adopts a unified network for both single-modal representation and multi-modal fusion, like VilBert \cite{NEURIPS2019_c74d97b0} and VLMo \cite{NEURIPS2022_d46662aa}. In the first line of work, the soft prompt method has been 
studied, like CoOp \cite{10.1007/s11263-022-01653-1}, CoCoOp \cite{Zhou_2022_CVPR}, UPT \cite{zang2022unified} and MaPLe \cite{Khattak_2023_CVPR}.
Nonetheless, there has been little work applying soft prompts in the second line of VLMs yet. Thus, exploring multi-modal prompts in a unified Transformer network remains an open issue.

In this paper, towards the deep MSU,
we propose a novel multi-modal soft prompt framework MoPE-BAF, \textbf{M}ixture-\textbf{o}f-\textbf{P}rompt-\textbf{E}xperts with \textbf{B}lock-\textbf{A}ware prompt \textbf{F}usion. Specifically, we devise a set of soft prompts corresponding with different roles, an image prompt expert, a text prompt expert and a unified prompt expert. The first two extract 
semantic features within a single modality, while the third assists in capturing inter-modality information. Furthermore, we introduce a block-aware prompt fusion mechanism to enhance the connection between different prompt experts. We re-organize the transformer layers into several blocks and apply cross-attention to enable the exchange of prompt expert information between two adjacent blocks. It facilitates deep interactions between modalities and enables smoother transitions from single-modal representation to multi-modal fusion.

We conduct a series of experiments on the MSDT dataset \citeplanguageresource{MSDT} in the few-shot setting. Our proposed model significantly outperforms the classical CLIP and the base VLMo. 
More importantly, our model with only 150M parameters obtains a better performance than InstructBLIP~\cite{dai2023instructblip},  the most advanced model with 8.2B parameters. 
Besides, experimental results demonstrate a stable performance gain compared to conventional soft prompts, no matter using a LM head or a classification head. Furthermore, we apply our model to the MSA task on the MVSA-S data \citeplanguageresource{MVSA}, outperforming the previous state-of-the-art UP-MPF \citep{10.1145/3503161.3548306} by 3.91 F1 points.   

To sum up, our contributions are: 
\begin{itemize}
    \item We propose a novel mixture-of-prompt-experts method on the unified VLMs, which drives the pre-trained model to get refined in both single-modal representation and multi-modal fusion. 
    \item We present a block-aware prompt fusion mechanism, which activates deep interactions between prompt experts and 
    balances the twin objectives of single-modal specialization and multi-modal fusion in VLMs. 
    \item We conduct experiments on the few-shot multi-modal sarcasm detection and sentiment analysis, outperforming the previous state-of-the-art methods and the advanced large language models.  
\end{itemize}

\section{Related Work}

\subsection{Multi-modal Sarcasm Detection}
Researchers have been exploring different methods to model the incongruity within the image-text pair for the MSD task. At the outset, feature-based approaches are adopted \cite{10.1145/2964284.2964321, castro-etal-2019-towards}. Subsequently, researchers pay more attention to modality interaction. 
\citet{cai-etal-2019-multi} leverages a hierarchical strategy to fuse the three modalities of image, attribute and text by attention weights. \citet{9206905} exploits the interaction among the input modalities using the recurrent neural network. \citet{pan-etal-2020-modeling} tries to capture the incongruity by inter-modal attention and co-attention within the text. \citet{xu-etal-2020-reasoning, 10.1145/3462244.3479919} decompose the model into a separation network for discrepancy increment and a relation network for relevance increment. 
\cite{10.1145/3474085.3475190, liang-etal-2022-multi, liu-etal-2022-towards-multi-modal} introduce graph structures for depicting incongruity relations.  
More recently, \citet{tian-etal-2023-dynamic} utilizes dynamic paths to activate different routing Transformers. However, these works rely on a large amount of training data to finetune the pre-trained models.  

\begin{figure*}[!ht]
    \centering
    \includegraphics[width=0.7\textwidth]{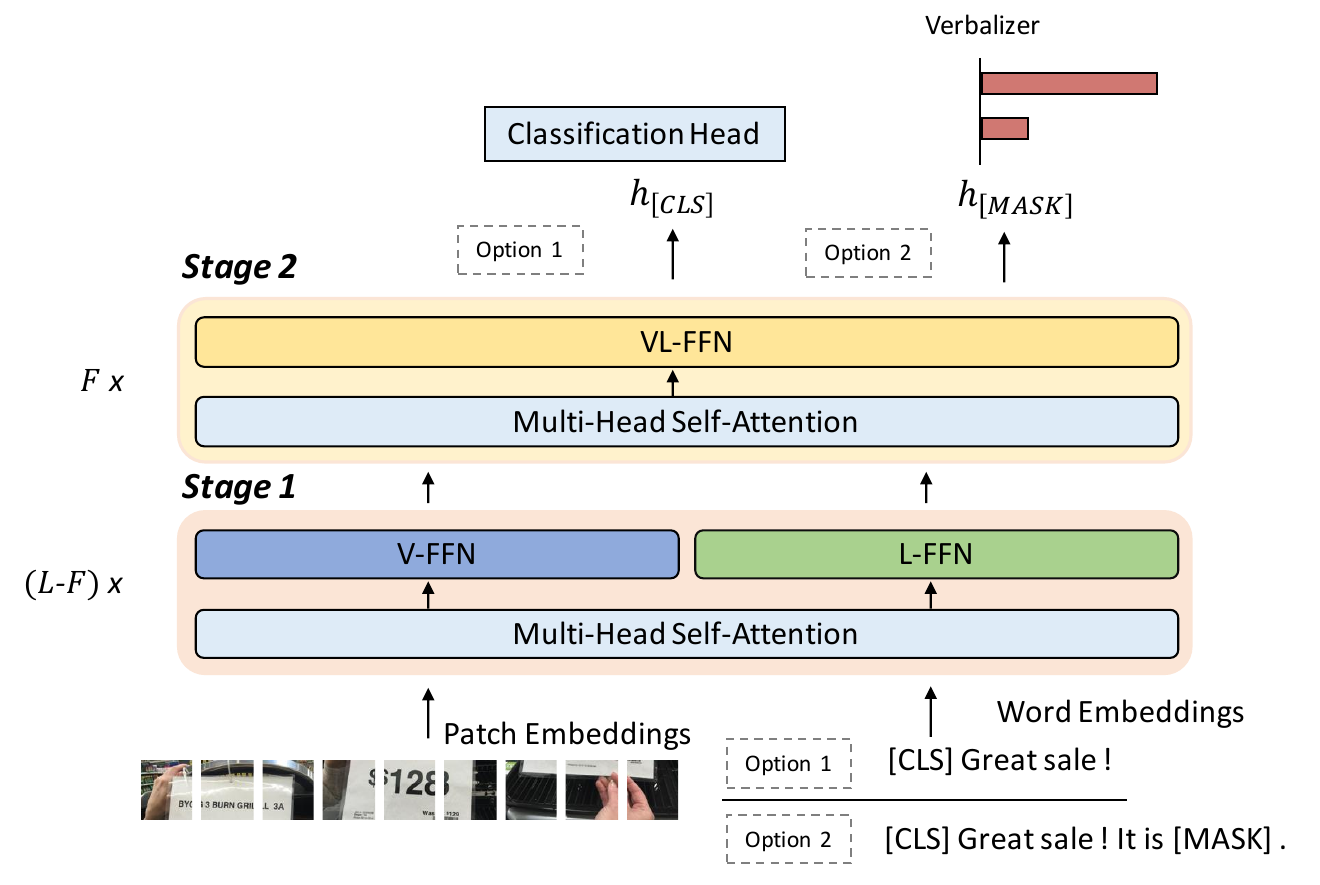}
    \caption{The framework of VLMo for image-text detection task. We demonstrate two methods.
    In \textit{finetuning}, the [CLS] representation is fed to a classification head,
    while in \textit{manual prompt}, the representation of [MASK] is fed to a verbalizer.}
    \label{fig:vlmo}
\end{figure*}

\subsection{Multi-modal Sentiment Analysis}
In recent years, 
MSA has become a popular research topic. 
\citet{10.1145/3209978.3210093} designs a co-memory network based on attention to predict the whole sentiment of text-image pairs. \citet{9246699} proposes a multi-view attention network, which utilizes an attention memory network 
to extract text and image features, and then fuses multi-modal features through a stacking-pooling module. After that, they incorporated graph neural network in MSA \citeyearpar{yang-etal-2021-multimodal}. \citet{10.1145/3503161.3548306} 
proposes a unified pre-training stage to narrow the semantic gap between different image and text pre-trained models.

\subsection{Multi-modal Prompt Learning}
Recently, researchers adapt VLMs to fit downstream tasks via prompt learning \cite{gu2023systematic}, which can be 
generally categorized 
into single-modal prompt and multi-modal prompt methods. In single-modal prompt methods, learnable continuous prompts are appended in front of the text data. 
For example, CoOp \cite{10.1007/s11263-022-01653-1} and CoCoOp \cite{Zhou_2022_CVPR} replace the manual-crafted text prompts used in CLIP 
with continuous vectors for image classification. Multi-modal prompt learning aims to optimize the text and image inputs simultaneously. 
UPT \cite{zang2022unified} 
designs a shared initial prompt for CLIP text and visual encoders.
MaPLe \cite{Khattak_2023_CVPR} leverages prefix tuning in both modality encoders and designs a coupling function to enable a mutual promotion between prompts. CMPA \cite{liu-etal-2023-deeply} 
designs prompts for both encoders and employ a cross-modal prompt attention at each layer.
These prompt-based methods are mainly applied to the CLIP architecture, which encodes the text and image inputs separately. To the best of our knowledge, there has been 
little work that applies soft prompts to a unified vision-language pre-trained model. 

\section{Preliminary}
Since our work adopts the vision-language pre-trained model VLMo as the backbone, to better illustrate our proposed method, 
we provide an overview of VLMo as well as the basic knowledge of applying it to MSU tasks.

\subsection{The Vision-language Pre-trained Model: VLMo}
VLMo presents 
a Transformer architecture with Mixture-of-Modality-Experts 
where the feed-forward neural (FFN) network switches based on the input modality and fusion requirements. As shown in Figure~\ref{fig:vlmo}, given an image-text pair, VLMo performs 
the unified encoding 
in two stages: (1) Employing vision FFN (V-FFN) and language FFN (L-FFN) to encode the respective modality representations at the bottom Transformer layers, and (2) Using vision-language FFN (VL-FFN) 
to 
perform multi-modal interaction at the top layers. 

Concretely, in Stage 1, denoting the hidden vision and language representations of the previous layer as $H_v^{n-1}, H_l^{n-1}$ respectively, VLMo first employs shared self-attention across modalities to align their contents, and then pass them 
to a uni-modal FFN to obtain the output of this layer:

\scriptsize
\begin{gather}
    H^{n-1} = \operatorname{concat}(H_v^{n-1}, H_l^{n-1}), \\
    \tilde{H}^n = \operatorname{softmax}(\frac{(H^{n-1}W_q)(H^{n-1}W_k)^\top}{\sqrt{d}})(H^{n-1}W_v), \\
    [\tilde{H}_v^n, \tilde{H}_l^n] = \tilde{H}^n \\
    H_v^n = \operatorname{V-FFN}(\tilde{H}_v^n), \\
    H_l^n = \operatorname{L-FFN}(\tilde{H}_l^n)
\end{gather}
\normalsize
In Stage 2, after the self-attention operation, the intermediate outputs are combined and forwarded to a vision-language FFN:
\begin{align}
    H_v^n, H_l^n = \operatorname{VL-FFN}([\tilde{H}_v^n, \tilde{H}_l^n]).
\end{align}

\subsection{Tuning on Multi-modal Tasks}
In MSD or MSA task, the input consists of an image $x$ and associated text $y$, with the goal of predicting its corresponding category.

\paragraph{Finetuning.} The most straightforward approach is to register a classification head on top of VLMo. After encoding, the vector of the text-start token ([CLS]) is used as the final representation of the image-text pair, and the prediction result is obtained:
\begin{align}
    p_{t} = \mathbb{P}(t|\theta, \phi, x, y),
\end{align}
where $\theta, \phi$ denote the parameters of the pre-trained model and the classification head respectively.

\paragraph{Manual Prompt.} Another way to utilize VLMo is to reformulate the task into a mask language modeling task using a hard template containing [MASK]:
\begin{align}
    p_t = \mathbb{P}(\textrm{[MASK]}=v(t)|\theta, T(x, y)),
\end{align}
where $T$ denotes the prompt template and $v$ denotes the verbalizer that maps the 
[MASK] token to the probabilities on label words.

\paragraph{Soft Prompt.} The soft prompt method utilizes a set of trainable virtual tokens, which eliminates the need to manually design templates:
\begin{align}
    T(x, y) = \textrm{[}V_1, V_2, \cdots, V_n\textrm{]}\ x\ ||\ y.
\end{align}
where $\{V_i\}$ is 
the virtual token.

\section{The Proposed Model}
\begin{figure*}
    \centering
    \includegraphics[width=\textwidth]{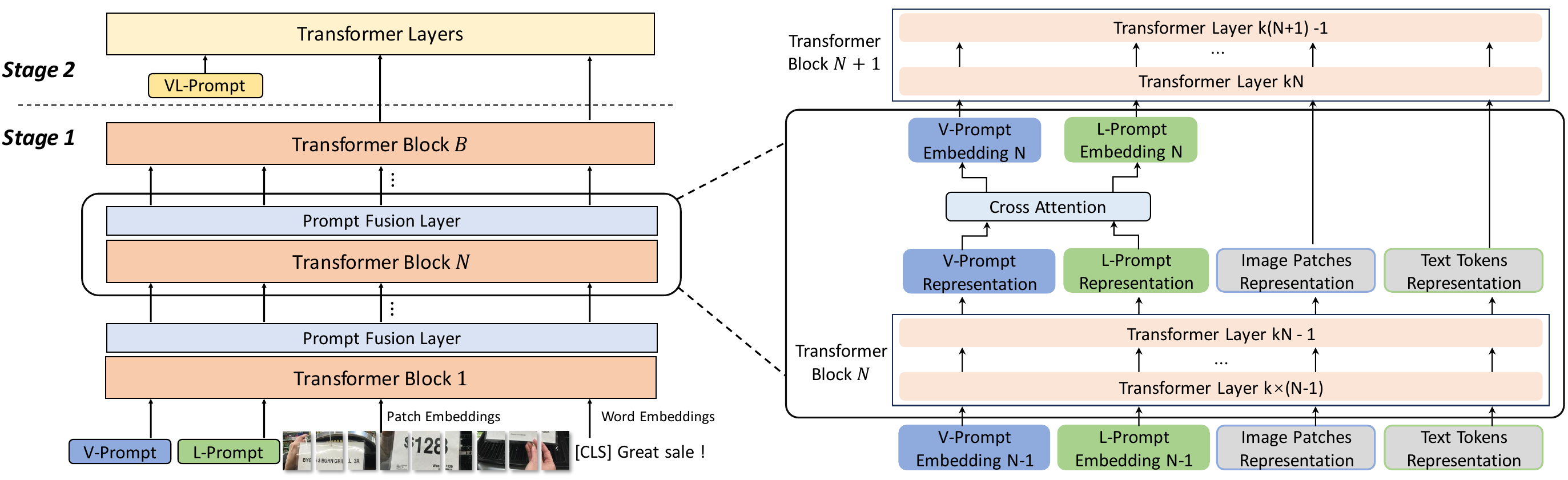}
    \caption{
    Our proposed MoPE-BAF model for multi-modal semantic understanding .}
    \label{fig:full-model}
\end{figure*}

\begin{figure}[!ht]
    \centering
    \includegraphics[width=0.48\textwidth]{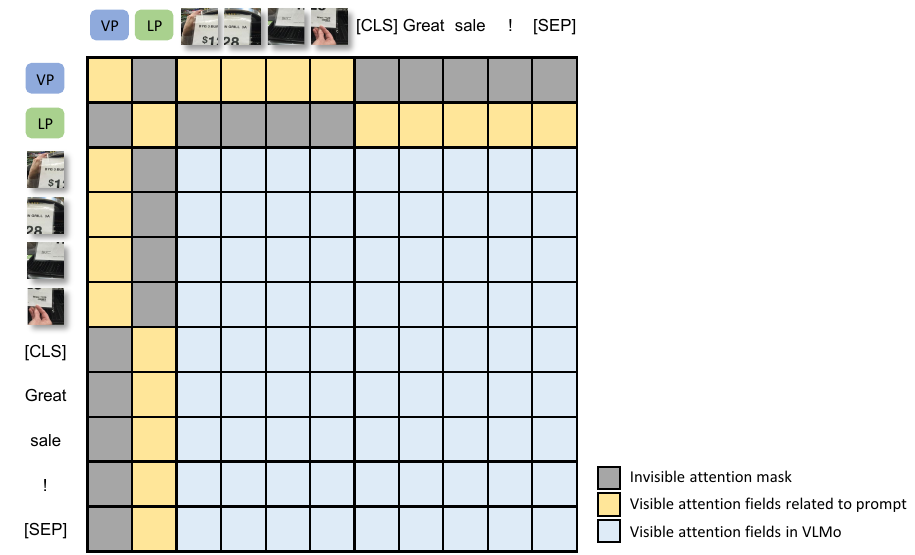}
    \caption{Receptive fields of different prompts, image patches, text tokens in the self-attention module when using MoPE. VP and LP are shorthand for V-Prompt, L-Prompt.}
    \label{fig:prompt attention}
\end{figure}

\subsection{Mixture-of-Prompt-Experts}
One of the primary challenges in applying soft prompts to multi-modal tasks 
is the specialization of prompt properties across different modalities. 
In the traditional soft prompt method, all prompts and input tokens are treated equally within Transformer layers, which aligns with the VLMo's second stage. However, VLMo differentiates image and text inputs with distinct FFNs in the first stage. Thus, only the second stage can be activated by prompts, which inhibits the full exploitation of the multi-modal encoder in extracting modality-specific features.

In view of this, we propose a novel multi-modal soft prompt approach, Mixture-of-Prompt-Experts (MoPE), to serve the two encoding stages in VLMo. 
It contains a set of soft prompts corresponding to the three functional FFNs of VLMo: image expert (V-Prompt), text expert (L-Prompt) and unified cross-modal expert (VL-Prompt), where V-Prompt and L-Prompt assist in extracting the semantic features from the respective modality in Stage 1, and VL-Prompt assists in enhancing inter-modal interaction. Each prompt expert is initialized as a set of trainable vectors with dimension matching the embeddings of the pretrained model. The overall structure of our proposed model is illustrated in Figure~\ref{fig:full-model}.

\paragraph{Stage 1.} We prepend a single-modal prompt to each modality input and pass them together through 
VLMo Transformer layers.

\scriptsize
\begin{gather}
    H^{n-1} = \operatorname{concat}(H_{vp}^{n-1}, H_{lp}^{n-1}, H_v^{n-1}, H_l^{n-1}), \\
    \tilde{H}^n = \operatorname{softmax}(\frac{(H^{n-1}W_q)(H^{n-1}W_k)^\top}{\sqrt{d}})(H^{n-1}W_v), \\
    [\tilde{H}_{vp}^n, \tilde{H}_{lp}^n, \tilde{H}_v^n, \tilde{H}_l^n] = \tilde{H}^n \\
    [H_{vp}^n, H_v^n] = \operatorname{V-FFN}([\tilde{H}_{vp}^n,\tilde{H}_v^n]), \\
    [H_{lp}^n, H_l^n] = \operatorname{L-FFN}([\tilde{H}_{lp}^n,\tilde{H}_l^n])
\end{gather}
\normalsize
where $H_{v}, H_{l}, H_{vp}, H_{lp}$ denote vision representation, language representation, V-Prompt and L-prompt, respectively. 

To ensure the specialization of prompt experts within their corresponding modalities, we restrict their receptive field in the self-attention module, as illustrated in Figure~\ref{fig:prompt attention}. V-Prompt is dedicated to the image input only, while the image can attend to both V-Prompt and the text input, enjoying the cross-modal alignment and uni-modal enhancement simultaneously. L-Prompt performs in the same way.

\paragraph{Stage 2.} A multi-modal unified prompt expert is 
introduced to enhance the interaction between modalities:

\scriptsize
\begin{gather}
    H^{n-1} = \operatorname{concat}(H_{vlp}^{n-1}, H_v^{n-1}, H_l^{n-1}), \\
    \tilde{H}^n = \operatorname{softmax}(\frac{(H^{n-1}W_q)(H^{n-1}W_k)^\top}{\sqrt{d}})(H^{n-1}W_v), \\
    [\tilde{H}_{vlp}^n, \tilde{H}_v^n, \tilde{H}_l^n] = \tilde{H}^n, \\
    [H_{vlp}^n, H_v^n, H_l^n] = \operatorname{VL-FFN}([\tilde{H}_{vlp}^n, \tilde{H}_v^n, \tilde{H}_l^n]),
\end{gather}
\normalsize
where $H_{vlp}$ is the representation of VL-Prompt.

\subsection{Block-Aware Prompt Fusion}
Recall that deep MSU
necessitates the ability to discern the complex relationships across image and text modalities. These tasks demand a profound understanding of content relations beyond simple fusion.
Despite VLMs being pre-trained on large image-text corpora and demonstrating strong performance on traditional multi-modal tasks (e.g., image classification),
their performance on the complex multi-modal task remains subpar due to the superficial fusion process.
In VLMo, the allocation of fusion layers is limited (e.g., VLMo-Base-plus assigns only 3 layers), 
constraining the model's fusion capacity.
Accordingly, we propose a new \textbf{b}lock-\textbf{a}ware prompt \textbf{f}usion (BAF) mechanism to make different modality prompts interact deeply and meet the fusion requirements in deep MSU.


To be specific, we re-organize the transformer layers 
into several blocks, and 
introduce a cross-attention fusion layer between two adjacent blocks. Assuming that each block contains $m$ layers, for the first layer of block $b$ (Layer $bm$), the input prompt 
is reconstructed from the output of the last layer of block $b-1$ (Layer $bm-1$):

\scriptsize
\begin{align}
    S_{vp}^{bm} = \operatorname{softmax}(\frac{(H_{lp}^{bm-1}W_q)(H_{vp}^{bm-1}W_k)^\top}{\sqrt{d}})(H_{vp}^{bm-1}W_v), \\
    S_{lp}^{bm} = \operatorname{softmax}(\frac{(H_{vp}^{bm-1}W_q)(H_{lp}^{bm-1}W_k)^\top}{\sqrt{d}})(H_{lp}^{bm-1}W_v),
\end{align}
\normalsize
where $S_{vp}^{bm}$ and $S_{lp}^{bm}$ are the input for Layer $bm$. 

Intuitively, an efficient prompt fusion should introduce knowledge from another modality while retaining the original specialization. We manage it through controlling the number of blocks in BAF. With an appropriate block number, the representations of different modality get fused gradually as the blocks progress, facilitating a seamless transition between two stages 
and striking a balance between single-modal specialization and multi-modal fusion.


\section{Experimental Setup}
\subsection{Dataset and Evaluation Metrics}
We conduct experiments on two representative MSU
tasks: multi-modal sarcasm detection and multi-modal sentiment analysis.

\paragraph{Multi-modal Sarcasm detection} For image-text MSD, the MSDT dataset \citeplanguageresource{MSDT} stands as the sole benchmark dataset currently available. MSDT has 29k/2.4k/2.4k sample pairs for train/validation/test, each of which contains an image-text pair with a binary label \{sarcasm, nonsarcasm\}. We keep the test set unchanged and randomly select 32 samples from the train/validation set to construct our few-shot dataset. To balance the label distribution, we control the samples for each label to account for half of the total number. Following the previous work, we adopt Accuracy and F1 as our evaluation metrics. To improve measuring robustness, we sample three disjoint datasets and report the mean result on them. The statistics of MSDT dataset is shown in Table \ref{tab:msdt-stats}.

\begin{table*}[!ht]
\tablefontsize
\centering
\begin{tabular}{@{}cl|cc|cc|c@{}}
\toprule
\multicolumn{2}{l|}{\multirow{2}{*}{}} & \multicolumn{2}{c|}{Full Split} & \multicolumn{2}{c|}{Few-shot Split} & \multirow{2}{*}{Avg. length} \\
\multicolumn{2}{l|}{} & \#Sarcasm & \#Nonsarcasm & \#Sarcasm & \#Nonsarcasm & \\ \midrule
\multicolumn{2}{c|}{Train} & 8642 & 11174 & 16 & 16 & 21.85 \\
\multicolumn{2}{c|}{Dev} & 959 & 1451 & 16 & 16 & 21.79 \\
\multicolumn{2}{c|}{Test} & 959 & 1450 & 959 & 1450 & 22.22 \\ \bottomrule
\end{tabular}
\caption{Statistics of MSDT dataset, with Avg. length calculated using Bert Tokenizer.}
\label{tab:msdt-stats}
\end{table*}

\paragraph{Multi-modal Sentiment Analysis} 
Our experiments for MSA are based on the MVSA-S \citeplanguageresource{MVSA} dataset. Each sample in MVSA-S contains an image-text pair annotated with one of the labels: \{positive, neutral, negative\}.
For few-shot MSA, \citet{10.1145/3503161.3548306} performed random sampling on both training and development sets from MVSA-S, constituting 1\% of the total. We keep the same setting with them for fair comparison and use Accuracy, Weighted-F1, and Macro-F1 as metrics.

\subsection{Implementation Details} \label{implementation-detail}
We choose VLMo-Base-plus as our baseline model. For data pre-processing, we follow the same steps as VLMo. The images are resized to $224\times 224$ resolution and segmented into $16 \times 16$ patches with RandAugment \citep{NEURIPS2020_d85b63ef}. We utilize the tokenizer from the uncased version of BERT, limiting the input text to a length of 40 and truncating any exceeding portions. For hyper-parameters, the block number is set to 2, and the prompt length is set to 10 by default. All experiments adhere to the full parameter training paradigm. During training, the model is optimized by AdamW \citep{loshchilov2018decoupled} with $\beta_1=0.9, \beta_2=0.998$. The peak learning rate is $3e-5$. Weight decay is 0.01. In the 32-shot setting, the batch size is 8, and we use linear warmup over the first 10\% of the whole 200 steps. 
We use 1 Nvidia GeForce 3080Ti card for experiments. One training needs around 8GB GPU memory and takes about 2 hours.

\begin{table*}[htbp]
\centering
\scriptsize
\begin{tabular}{@{}l|l|l@{}}
\toprule
 & \textbf{MSD} & \textbf{MSA} \\ \midrule
Manual Prompt & The image-text pair is [MASK]. <text> & Sentiment of the text: [MASK]. <text> \\
Soft Prompt & [V1] [V2] ... [Vn] <text> & [V1] [V2] ... [Vn] <text> \\
P-Tuning & [V1] [V2] ... [Vn] The image-text pair is [MASK]. <text> & [V1] [V2] ... [Vn] Sentiment of the text: [MASK]. <text> \\ \midrule
\multirow{8}{*}{Instruction Prompt} & \multicolumn{1}{p{0.4\textwidth}|}{1.Text:<text> Answer the question: Is this image-text pair sarcastic or nonsarcastic? Answer:\newline 2.Text:<text> Based on the image and text, answer the question: Is this image-text pair sarcastic or nonsarcastic? Answer: \newline
3.Text: <text> Combining the text, is this sarcastic or nonsarcastic? Answer:} & \multicolumn{1}{p{0.4\textwidth}}{1.Text:<text> Answer the question: Which sentiment does this image-text pair contain, negative, neutral or positive? Answer:\newline 2.Text:<text> Based on the image and text, answer the question: Which sentiment does this image-text pair contain, negative, neutral or positive? Answer:\newline 3.Text:<text> Combining the text, which sentiment does this contain, negative, neutral or positive? Answer:} \\ \bottomrule
\end{tabular}
\caption{The text templates used in different prompt methods for MSD and MSA tasks. The instruction prompt is for InstructBLIP. We do not include P-Tuning v2 here as it does not require a template for input texts.}
\label{tab:prompt-design}
\end{table*}

\subsection{Comparing Methods}
We conduct a comprehensive evaluation by comparing with the vanilla multi-modal pre-trained methods, the methods specialized for MSD or MSA tasks, and the prompt methods applied to VLMs.
\subsubsection*{Multi-modal Pre-trained methods}
\textbf{CLIP} \citep{pmlr-v139-radford21a} holds a dual-encoder architecture, designed to encode texts and images separately.
We further add trainable projection layers at the output of both encoders and perform classification on their concatenated encoding results. \textbf{InstructBLIP} \citep{dai2023instructblip} is the latest and most advanced general-purpose VLM that introduces instruction tuning techniques to extract informative features tailored to the given instruction. We adopt the Vicuna7B as the base and design three textual prompts for MSD and MSA as the instruction (shown in Table \ref{tab:prompt-design}) and report their average performance. We use 1 Nvidia A40 card for training InstructBLIP and it takes around 24GB GPU memory.


\subsubsection*{Multi-modal Prompt Learning Methods}
We also consider the multi-modal prompt methods based on CLIP as our baselines. \textbf{CoOp} \citep{10.1007/s11263-022-01653-1} replaces the manual-crafted text prompts in CLIP with continuous vectors. \textbf{MaPLe} \citep{Khattak_2023_CVPR} leverages prefix tuning in both modality encoders and designs a coupling function to enable prompt interactions. \textbf{CMPA} \citep{liu-etal-2023-deeply} designs prompts for both encoders and employs a cross-modal prompt attention at each layer.

\subsubsection*{Multi-modal Sarcasm Detection methods}
For the MSD task, previous work has only considered a full-shot experimental setup. Based on our few-shot dataset, we replicate some of the studies for which the source code is available. 
\textbf{HFM} \citep{cai-etal-2019-multi} leverages a hierarchical strategy to fuse the image, attribute and text modalities by attention weights. \textbf{ResBERT} \citep{pan-etal-2020-modeling} captures the cross-modal incongruity by inter-modal attention and contradiction within the text by co-attention. \textbf{HKE} \citep{liu-etal-2022-towards-multi-modal} learns composition-level congruity based on graph neural networks and introduces auto-generated captions as external knowledge. 

\subsubsection*{Multi-modal Sentiment Analysis methods}
\textbf{MVAN} \cite{9246699} 
utilizes a multi-view memory network to extract single-modal emotion features and interactively capture the cross-view dependencies between the image and text. \textbf{MGNNS} \cite{yang-etal-2021-multimodal} learns multi-modal representations by a multi-channel graph neural network based on the global characteristics of MVSA. 
\textbf{UP-MPF} \cite{10.1145/3503161.3548306} adopts the multi-modal prompt-based finetuning paradigm and proposes a pre-training stage to narrow the semantic gap between image and text encoders.

\subsubsection*{Different Prompting Learning Strategies}
\textbf{Manual Prompt} defines a hard template combined with input. The representation of [MASK] is then fed into the verbalizer for prediction. \textbf{Soft Prompt} prepends several virtual tokens before the input and utilizes [CLS] token for classification.
\textbf{P-Tuning} \cite{liu2021gpt} combines soft prompt with manual prompt. Table \ref{tab:prompt-design} lists the templates used in these methods. \textbf{P-Tuning v2} \cite{DBLP:journals/corr/abs-2110-07602} adopts the prefix-tuning idea and expands the prompt parameter space to each transformer layer, using [CLS] token for classification. 


\begin{table*}[htbp]
\centering
\tablefontsize
\begin{tabular}{@{}lcccccc@{}}
\toprule
\textbf{Methods} & \multicolumn{1}{c}{\textbf{w/ CH}} & \multicolumn{1}{c}{\textbf{w/ LMH}} & \textbf{Accuracy} & \textbf{Precision} & \textbf{Recall} & \textbf{F1} \\ \midrule
HFM \citep{cai-etal-2019-multi} & \checkmark &  & 58.88 (2.93) & 48.93 (2.00) & 65.83 (12.28) & 55.73 (4.51) \\
resBERT \citep{pan-etal-2020-modeling} & \checkmark &  & 58.85 (2.04) & 48.95 (1.62) & 70.87 (4.61) & 57.82 (0.45) \\
HKE \citep{liu-etal-2022-towards-multi-modal} & \checkmark &  & 60.76 (4.00) & 52.56 (3.41) & 74.43 (14.13) & 61.02 (3.80) \\ \midrule
VLMo \citep{NEURIPS2022_d46662aa} & \checkmark &  & 60.18 (3.62) & 50.97 (4.26) & 60.31 (5.28) & 53.96 (3.65) \\
CLIP \citep{pmlr-v139-radford21a} & \checkmark &  & 61.31 (1.28) & 51.24 (1.30) & 68.33 (8.26) & 58.26 (2.22) \\
InstructBLIP (zero-shot) &  & \checkmark & 57.59 (2.87) & 47.22 (3.12) & 50.33 (13.25) & 47.89 (5.49) \\
InstructBLIP \citep{dai2023instructblip} &  & \checkmark & 60.56 (4.02) & 50.57 (3.29) & \textbf{78.13} (5.94) & 61.05 (2.11) \\ \midrule
CoOp \citep{10.1007/s11263-022-01653-1} & \checkmark &  & 63.00 (5.99) & 53.06 (6.42) & 67.71 (4.69) & 59.41 (5.23) \\
CMPA \citep{liu-etal-2023-deeply} & \checkmark &  & 59.94 (2.34) & 49.75 (2.35) & 63.43 (4.08) & 55.75 (2.86) \\
MaPLE \citep{Khattak_2023_CVPR} & \checkmark &  & 61.28 (2.43) & 50.87 (1.95) & 78.00 (6.42) & 61.26 (2.80) \\
\midrule
VLMo + Manual Prompt &  & \checkmark & 59.85 (2.77) & 50.12 (2.64) & 73.20 (8.25) & 59.10 (1.15) \\
\quad + Soft Prompt & \checkmark &  & 62.74 (0.65) & 53.26 (1.56) & 57.84 (11.09) & 54.74 (4.71) \\
\quad + P-Tuning \citep{liu2021gpt} &  & \checkmark & 60.34 (1.61) & 50.11 (1.31) & 75.46 (4.03) & 60.21 (2.06) \\
\quad + P-Tuning v2 \citep{DBLP:journals/corr/abs-2110-07602} & \checkmark &  & 61.78 (3.02) & 52.23 (2.94) & 63.99 (15.12) & 56.36 (5.98) \\ \midrule
VLMo + MoPE-BAF & \checkmark &  & 64.06 (0.71) & 53.69 (4.87) & 71.60 (2.78) & 61.32 (2.57) \\
\qqquad + MoPE-BAF + MP &  & \checkmark & \textbf{65.32} (3.30) & \textbf{55.92} (3.91) & 68.64 (7.52) & \textbf{61.32} (1.15) \\ \bottomrule
\end{tabular}
\caption{Experimental results on 
the MSDT dataset with 32 training samples, where the standard deviations are shown in parentheses. CH refers to the classification head and LMH refers to the language modeling head. MP means Manual Prompt. }
\label{tab:main-results}
\end{table*}

\section{Results and Analysis}

\subsection{Evaluation on Multi-modal Sarcasm Detection}
We conduct extensive experiments on the dataset MSDT, and report the results in Table \ref{tab:main-results}. We implement two settings for our method. The first employs a classification head using [CLS] for prediction, termed MoPE-BAF. The second uses an additional manual template containing [MASK], termed MoPE-BAF + Manual Prompt (our full model).

Compared to previous methods that target full-shot MSD, our model MoPE-BAF achieves significant improvements over HFM and resBERT, and surpasses HKE by a margin of 4.56 points on Accuracy, which leverages external caption knowledge.


Among the current advanced multi-modal models, our method achieves promising results. Compared with the backbone model VLMo, we realize 5.14 and 7.36 points improvement on Accuracy and F1, respectively. Compared with CLIP, our methods attain consistent improvements across all four metrics. In addition, our model with only 150m parameters surpasses the large model InstructBLIP with 8.2B parameters, by 4.76 points increment on Accuracy. This efficiency highlights the superiority of our methods and suggests our potential for practical resource-saving applications. 

In comparison to the multi-modal prompt learning methods CoOp, MaPLE and CMPA that implemented on CLIP, MoPE-BAF showcases outstanding performance despite being built on a less powerful backbone model VLMo. Besides, under the scope of prompt methods on VLMo, our full model achieves the best results, whether combined with a classification head or a LM head with the verbalizer. We speculate that our method distinguishes prompts for different modalities and facilitates the interaction between them as the layers deepen, which enhances the specialization of single-modal representation and guarantees a thorough modality fusion. Thus, we obtain significant improvements compared with the non-specific prompt methods P-Tuning and P-Tuning v2, even though P-Tuning v2 requires a larger prompt parameter space for prepending soft prompts in each Transformer layer. 


\subsection{Evaluation on Multi-modal Sentiment Analysis}

\begin{table}
\tablefontsize
\centering
\begin{tabular}{@{}lccc@{}}
\toprule
\textbf{Methods} & \textbf{Acc}   & \textbf{Mac-F1} & \textbf{Wtd-F1} \\ \midrule
MVAN & 42.77 & 36.75  & 44.14  \\
MGNNS & 34.4  & 32.05  & 36.9   \\
UP-MPF & 58.21 & 51.08  & 58.49  \\ \midrule
CLIP & 49.51 & 45.67 & 51.63 \\
CoOp & 51.47 & 40.58 & 48.52 \\
MaPLe & 50.49 & 43.06 & 51.74 \\
CMPA & 56.74 & 42.75 & 53.86 \\
InstructBLIP & 59.80 & 48.59  & 59.73  \\
VLMo + MP (zero-shot) & 59.07 & 38.53 & 51.94 \\
VLMo + MP & 60.79 & 52.62 & 61.27 \\
VLMo + P-Tuning & 61.03 & 51.28 & 60.75 \\

VLMo + MoPE-BAF + MP & \textbf{63.48} & \textbf{52.92}  & \textbf{62.40}  \\ \bottomrule
\end{tabular}
\caption{Comparison of results between our approach and previous methods on the few-shot MVSA-S dataset.
Mac-F1 and Wtd-F1 denote Macro-F1 and Weighted-F1 respectively. MP denotes manual prompt. The results of the first group are from the work \citep{10.1145/3503161.3548306}, and we implemented the models in the second group.}
\label{tab:mvsa}
\end{table}

\begin{figure*}[htbp]
\centering
\includegraphics[width=0.9\textwidth]{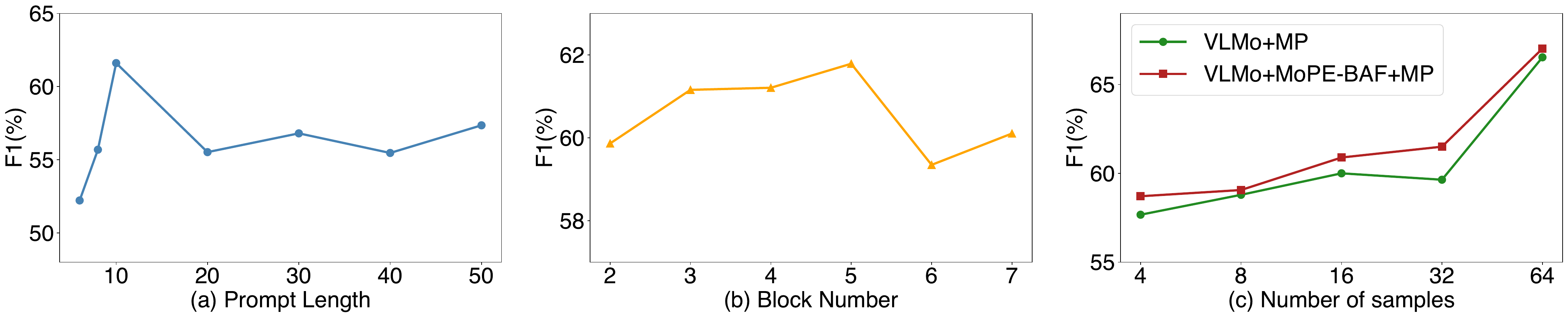}
\caption{(a) F1 performance 
    training 
    MoPE with different prompt lengths. (b) F1 
    scores training 
    MoPE-BAF with different block numbers. (c) Comparison between VLMo and VLMo + MoPE-BAF under different training shots.}
\label{fig:analysis}
\end{figure*}

For the MSA task, the experimental results on MVSA-S dataset are shown in Table \ref{tab:mvsa}.  
There is a huge performance gap between the non-pretrained methods (MVAN, MGNNS) and pre-trained methods. 
Among the pre-trained multi-modal methods, InstructBLIP demonstrates impressive efficacy, outperforming the current state-of-the-art UP-MPF. 
VLMo series perform better than CLIP series in this task, perhaps because the unified structure of VLMo, which effectively leverages both text and image input, is well-suited to this type of task, and the abundant pre-training tasks (especially mask language modeling), endows it with the ability to infer sentiment. 

VLMo performs well in the zero-shot setting, establishing a robust baseline. When integrated with prompts, the performance of VLMs is significantly boosted. VLMo adopting prompt-based finetuning (+MP) delivers the most superior results compared to those without VLMo, even outperforming the large-scale language model InstructBLIP. After combining with our MoPE-BAF method, the performance is further elevated, surpassing UP-MPF significantly by 3.91 F1 points. It demonstrates the generality and validity of our proposed model in different MSU tasks.

\subsection{Ablation Study}
\begin{table}
\tablefontsize
\centering
\begin{tabular}{@{}l|cccc@{}}
\toprule
\textbf{Methods} & \textbf{A} & \textbf{P} & \textbf{R} & \textbf{F1} \\ \midrule
VLMo & 60.18 & 50.97 & 60.31 & 53.96 \\
\qqquad + MoPE & 61.73 & 51.35 & \textbf{75.13} & 60.94 \\
\qqquad + MoPE + BAF & \textbf{64.06} & \textbf{53.69} & 71.60 & \textbf{61.32} \\ \midrule
VLMo + MP & 59.85 & 50.12 & \textbf{73.20} & 59.10 \\
\qqquad \quad + MoPE & 64.06 & 53.98 & 70.73 & 60.93 \\
\qqquad \quad + MoPE + BAF & \textbf{65.32} & \textbf{55.92} & 68.64 & \textbf{61.32} \\ \bottomrule
\end{tabular}
\caption{Ablative analysis of the MoPE and BAF modules.} 
\label{table:Ablation}
\end{table}
We conduct ablation experiments to investigate the effects of MoPE and BAF, 
and the results are presented in Table~\ref{table:Ablation}. 
In both settings (with or without using Manual Prompt), MoPE significantly improves the performance of VLMo, and further adding the prompt fusion operation achieves better performance.
Please note that we cannot validate the effect of BAF separately since it cannot exist independently of MoPE.

\subsection{Model Analysis}

We also conduct experiments to analyze some controllable factors in the proposed MoPE-BAF model. 

\paragraph{Prompt Length}
Generally, a longer prompt correlates with an increase in learnable parameters, while in the few-shot setting, it may exacerbate the over-fitting problem. We vary the prompt length from 5 to 50 in MoPE, and the results concerning different lengths are visualized in Figure~\ref{fig:analysis} (a). Overall, the impact of prompt length on model performance shows a trend of initially increasing, subsequently diminishing, and then stabilizing. Our explanation is that prompts too short are insufficient to bring about qualitative changes to the model, while too long may pose challenges for searching for the optimal solution due to the complexity of the expanded search space.

\paragraph{Block Number}
We investigate the impact of 
the block number in the BAF module. Specifically, in the first 21 layers of VLMo that use V-Prompt and L-Prompt experts, we divide them into 2-7 blocks. A configuration with 1 block implies no prompt fusion, and is therefore not displayed here. If the number of model layers cannot be evenly divided by the block number, 
the excess layers are allocated to the bottom blocks. This ensures that the maximum layer difference 
between any two blocks does not exceed 1. The results are shown in Figure~\ref{fig:analysis} (b). 
When the block number is between 3 and 5, the overall performance remains 
almost similar. However, a noticeable decline can be observed after 6 blocks. We speculate that too many cross-modal cross-attention operations between blocks cause their original specialization to be discarded, thereby contradicting the function of MoPE. An appropriate number of blocks can strike a balance between 
the specialization and different modalities interaction. 
\paragraph{Training Shots}
The number of training shots plays a crucial role in model performance, particularly in the few-shot setting. We conduct experiments to train the VLMo base and our full model with different numbers of samples. The results are presented in Figure~\ref{fig:analysis} (c). We find that the performance of both models improves as training shots increase. Besides, applying MoPE-BAF consistently improves the performance of VLMo with the number of training examples from 4 to 64 in the f1 score, which demonstrates the robustness and efficiency of our method.

\section{Conclusion}
In this paper, we present MoPE-BAF, a new multi-modal soft prompt framework catering to unified VLMs for few-shot multi-modal tasks. Specifically, we 
devise two 
prompt experts to serve the text and image modality separately
with a better specialization ability, and further activate the interactions of prompt experts by inserting cross-modal prompt attention between adjacent Transformer 
blocks.
In this way, we reach a harmonious balance in modality specialization and fusion, thus fulfilling the requirement of modeling deep relations between modalities. 
Experiments on multi-modal sarcasm detection and multi-modal sentiment analysis show that our MoPE-BAF model
not only surpasses other widely-used prompt methods, 
but also outperforms the advanced large language models. 
Further analysis confirms the effectiveness of MoPE and BAF.  

In the future, we intend to incorporate task-related external knowledge into our prompt design, and broaden the scope of our method to include other tasks, such as multi-modal content generation and multi-modal reasoning.

\section{Limitations}
While our study provides insights into the soft prompt technique on VLMs, it has to be acknowledged that it has some limitations. First, we observe a sensitivity of MoPE during training. The performance of MoPE relies on appropriate hyperparameter selection and the optimal hyperparameters differ across downstream tasks. While performing perform a grid search on hyperparameters may mitigate the issue, we believe that how to effectively control the sensitivity is worth further exploration in future work.

Besides, although MoPE-BAF can theoretically be applied to any unified VLMs without disrupting the architecture or encoding process of the base architecture, our study was conducted exclusively on VLMo, and we have not yet extended MoPE-BAF to other pre-trained VLMs. This restricts the generalizability analysis of our methods. In the future, we would apply MoPE-BAF on more VLMs, which we think would provide a more comprehensive understanding of the role and impact of our methods.

\nocite{*}
\section*{Acknowledgement}
This work is supported by the Key Project of Natural Science Foundation of China (61936012) and the National Natural Science Foundation of China (62076008).
\section*{Bibliographical References}\label{sec:reference}

\bibliographystyle{lrec-coling2024-natbib}
\bibliography{lrec-coling2024-example}

\section*{Language Resource References}
\label{lr:ref}
\bibliographystylelanguageresource{lrec-coling2024-natbib}

\bibliographylanguageresource{languageresource}

\end{document}